\documentclass[sigconf]{acmart}
\renewcommand\footnotetextcopyrightpermission[1]{}
\usepackage{graphicx}
\usepackage{stfloats}
\usepackage{wrapfig}
\usepackage{lipsum}  
\usepackage[capitalise]{cleveref}

\newlength\savewidth
\newcommand\shline{\noalign{\global\savewidth\arrayrulewidth
                            \global\arrayrulewidth 1.5pt}%
                   \hline
                   \noalign{\global\arrayrulewidth\savewidth}
                   }

\usepackage{xcolor}

\usepackage{algorithm}
\usepackage[noend]{algpseudocode}

\algrenewcommand\alglinenumber[1]{#1}

\usepackage{soul}

\usepackage{multirow}

\usepackage{enumitem}
\setlist[itemize]{leftmargin=*, topsep=.0em, itemsep=0pt, parsep=0pt, partopsep=0pt}
\setlist[enumerate]{leftmargin=*, topsep=.0em, itemsep=0pt, parsep=0pt, partopsep=0pt}

\usepackage[frozencache,cachedir=.]{minted}

\usepackage{amsthm}
\newtheoremstyle{def_style}
  {0.5em}     
  {0.5em}     
  {}          
  {}          
  {\bfseries} 
  {.}         
  {0.5em}      
  {}          

\theoremstyle{def_style}

\theoremstyle{def_style}

\usepackage{amsmath}
\usepackage{bbm}

\usepackage{collectbox}



\usepackage{mathtools}

\usepackage{balance}

\usepackage{caption}
\usepackage{subcaption}

\setcopyright{acmlicensed}
\copyrightyear{2018}
\acmYear{2018}
\acmDOI{XXXXXXX.XXXXXXX}
\acmConference[Conference acronym 'XX]{Make sure to enter the correct
  conference title from your rights confirmation email}{June 03--05,
  2018}{Woodstock, NY}
\acmISBN{978-1-4503-XXXX-X/2018/06}

\begin{document}

\newcommand{\proposed}{DCATS}

\title{Empowering Time Series Forecasting with LLM-Agents}









\author{Chin-Chia Michael Yeh}
\authornote{miyeh@visa.com}
\author{Vivian Lai}
\author{Uday Singh Saini}
\affiliation{%
  \institution{Visa Research}
  \city{Foster City}
  \state{CA}
  \country{USA}
}

\author{Xiran Fan}
\author{Yujie Fan}
\author{Junpeng Wang}
\affiliation{%
  \institution{Visa Research}
  \city{Foster City}
  \state{CA}
  \country{USA}
}

\author{Xin Dai}
\author{Yan Zheng}
\affiliation{%
  \institution{Visa Research}
  \city{Foster City}
  \state{CA}
  \country{USA}
}

\renewcommand{\shortauthors}{Chin-Chia Michael Yeh et al.}

\begin{abstract}
Large Language Model (LLM) powered agents have emerged as effective planners for Automated Machine Learning (AutoML) systems.
While most existing AutoML approaches focus on automating feature engineering and model architecture search, recent studies in time series forecasting suggest that lightweight models can often achieve state-of-the-art performance.
This observation led us to explore improving data quality, rather than model architecture, as a potentially fruitful direction for AutoML on time series data.
We propose \proposed{}, a \underline{D}ata-\underline{C}entric \underline{A}gent for \underline{T}ime \underline{S}eries.
\proposed{} leverages metadata accompanying time series to clean data while optimizing forecasting performance.
We evaluated \proposed{} using four time series forecasting models on a large-scale traffic volume forecasting dataset.
Results demonstrate that \proposed{} achieves an average 6\% error reduction across all tested models and time horizons, highlighting the potential of data-centric approaches in AutoML for time series forecasting.
The source code is available at: \url{https://sites.google.com/view/ts-agent}.
\end{abstract}




\maketitle

\section{Introduction}
Time series data pervades diverse real-world applications, including traffic monitoring, financial transactions, and ride-share demand forecasting~\cite{liu2023largest}.
Accurate time series forecasting, in particular, is a critical research area with extensive applications~\cite{godahewa2021monash,zeng2023transformers,liu2023largest}.
Concurrently, Large Language Models (LLMs) have demonstrated considerable success in powering Automatic Machine Learning (AutoML) systems for various general machine learning tasks~\cite{huang2023mlagentbench,wang2024openhands,schmidt2024aide,chan2024mle}.
This confluence motivates our exploration into developing an LLM-powered AutoML system specifically designed for the unique challenges of time series forecasting.

However, directly applying general AutoML principles to time series forecasting can overlook domain-specific optimization opportunities.
Recent findings indicate that lightweight time series models can achieve state-of-the-art performance~\cite{zeng2023transformers,chen2023tsmixer,lin2024sparsetsf,yeh2025compact}, shifting focus towards data quality.
This aligns with the principles of data-centric AI~\cite{wang2024visual,zha2025data}, which prioritizes data refinement over complex model architectures.
We identify a significant opportunity in leveraging LLMs to automate this data-centric approach for time series forecasting, an area that remains underexplored.
Therefore, this paper introduces \proposed{} (\underline{D}ata-\underline{C}entric \underline{A}gent for \underline{T}ime \underline{S}eries), an LLM-powered agent that focuses on intelligently refining training data rather than solely optimizing model architectures.

\begin{figure}[htp]
\centerline{
\includegraphics[width=0.9\linewidth]{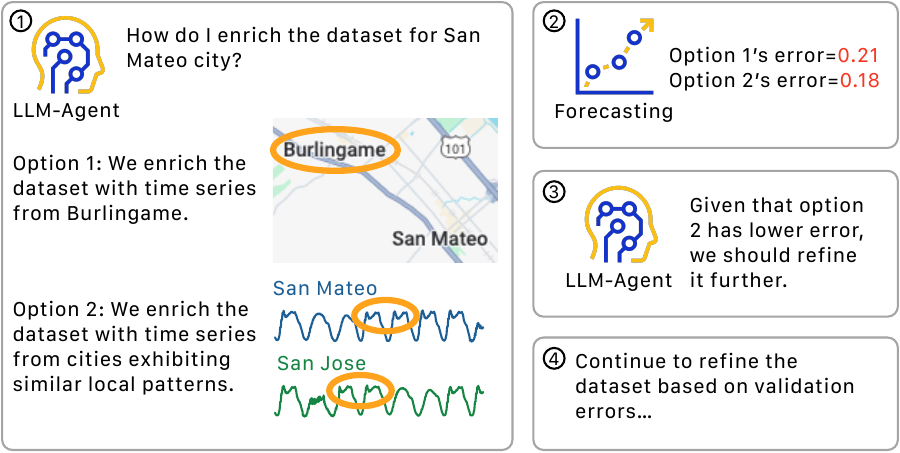}
}
\caption{
The proposed \proposed{} framework focuses on refining data quality rather than model design. 
Note, the LLM-agent makes decisions based on validation errors.
}
\label{fig:motivation}
\end{figure}

\proposed{} operates by strategically enriching the training dataset through the selection of relevant auxiliary time series.
The LLM-agent formulates a dataset expansion plan by reasoning over the metadata associated with available time series.
For instance, as illustrated in \cref{fig:motivation}, to forecast traffic volume for a highway entrance near San Mateo, California, \proposed{} might identify and incorporate historical data from proximate locations like Burlingame or from geographically distant areas exhibiting similar temporal patterns.
Furthermore, the agent iteratively refines this data selection plan by evaluating the impact of different enrichment strategies, thereby optimizing the final dataset.

To validate our approach, we conducted a preliminary study employing a large-scale traffic volume forecasting dataset~\cite{liu2023largest}, notable for its extensive metadata.
Our findings demonstrate that \proposed{} achieves a 6\% reduction in forecasting error compared to a baseline model trained using all available time series.
This result underscores the efficacy of leveraging LLM-agents for data-centric optimization in time series forecasting and signals a promising avenue for future research across various domains.

Our key contributions are as follows:
\begin{itemize}
    \item We introduce \proposed{}, a novel data-centric agentic framework designed specifically for time series forecasting problems.
    \item We present a preliminary study using traffic volume time series, showcasing a 6\% performance improvement over alternative solutions that do not employ LLM-agents.
    \item We showcase the ability of LLMs to perform reasoning over time series metadata to formulate effective data enrichment strategies for improved forecasting accuracy.
\end{itemize}

\section{Related Work}
Traditional AutoML systems primarily focus on model selection, hyperparameter tuning, and pipeline configuration~\cite{thornton2013auto,olson2016tpot,feurer2022auto}, employing various optimization and meta-learning techniques to achieve automation.
A parallel development has been neural architecture search, which specifically targets the automated design of neural networks using reinforcement learning or evolutionary algorithms~\cite{zoph2016neural,liu2018darts,pham2018efficient,real2019regularized,elsken2019neural}.
However, these approaches predominantly emphasize model optimization rather than data quality enhancement.
Recent advancements in LLMs have enabled more sophisticated AutoML approaches, with systems like AIDE~\cite{schmidt2024aide} employing LLM-agents for solving general machine learning problems.
While AIDE demonstrates the potential of LLM-powered AutoML, it does not fully exploit the unique characteristics of time series data.
The emerging paradigm of data-centric AI suggests that improving data quality can yield greater performance gains than model refinement alone~\cite{wang2024visual,zha2025data}, particularly relevant for time series forecasting where recent research shows lightweight models with high-quality data can match or exceed complex architectures~\cite{zeng2023transformers,chen2023tsmixer,lin2024sparsetsf,yeh2025compact}.
Despite these findings, few AutoML systems have been designed with a data-centric focus for time series forecasting, which presents unique challenges including temporal dependencies and domain-specific metadata~\cite{godahewa2021monash}.
To the best of our knowledge, \proposed{} represents the first LLM-powered data-centric AutoML framework designed specifically for time series forecasting problems, leveraging rich metadata to intelligently select and augment training data.

\section{Methodology}
\begin{figure}[htp]
\centerline{
\includegraphics[width=0.8\linewidth]{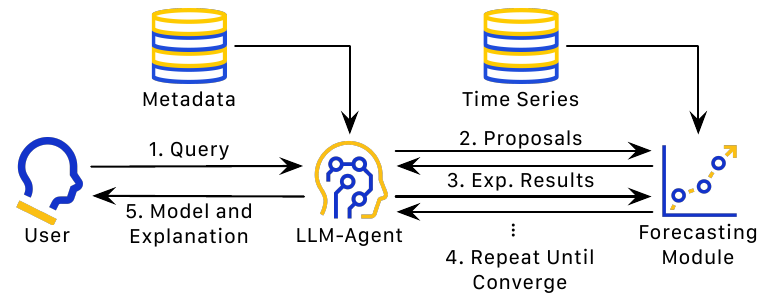}
}
\caption{
Overall design of the proposed \proposed{}.
}
\label{fig:overall}
\end{figure}

The proposed \proposed{} framework comprises four core components, as illustrated in \cref{fig:overall}.
We will use the traffic volume prediction use case to demonstrate the application of the proposed framework for time series forecasting.
The key components of \proposed{} are:
1) a time series dataset consisting of multiple univariate time series,
2) an accompanying metadata database containing shared background information and additional details about each univariate time series,
3) an LLM-agent responsible for driving the automatic forecasting model building process, and
4) a forecasting module that builds models and validates performance based on the agent's requests.

The framework is initiated when users submit a query expressing their intent to build a forecasting model for specific time series within the dataset.
Upon receiving the query, the LLM-agent retrieves background information about the entire time series dataset and detailed information about each time series from the metadata database.
Based on this information, the agent generates a list of proposals, each containing instructions for creating a sub-dataset from the time series dataset and an explanation of the strategy behind it.
The proposal generation process is detailed in \cref{sec:init_proposal}.

Each proposal is then evaluated by the forecasting module, which trains models using the sub-dataset and reports performance on the validation dataset.
The forecasting module is described in detail in \cref{sec:forcasting}.
Using the validation performance, the LLM-agent provides a new set of proposals for the next evaluation round, typically refining the winning strategy from the previous iteration.
The refinement process is explained in \cref{sec:refinement}.
This iterative process continues until no proposals in the current round show improvement over the best-so-far solution.

\subsection{Initial Proposal}
\label{sec:init_proposal}
The LLM-agent generates initial proposals using a five-section prompt:
1) background, 2) task, 3) guidelines, 4) neighbor sets, and 5) output format.
The background section provides context for the time series dataset.
For our traffic volume forecasting problem, the prompt is shown in \cref{list:init_background}.

\begin{listing}[htp]
\begin{minted}[breaklines,xleftmargin=2em,linenos,fontsize=\scriptsize, escapeinside=!!,breaksymbolleft=]{markdown}
# Background
We have a spatio-temporal dataset containing 8,600 locations, each with a unique `location_id` (integer from 0 to 8,599). Each location has a univariate time series representing traffic volume changes over time, split into training and validation datasets. Our goal is to build time series forecasting models for specific locations.
\end{minted}
\caption{The background.}
\label{list:init_background}
\end{listing}

The task section includes information about the target time series and the specific task (i.e., providing a number of proposals).
An example task section for location 1201 is shown in \cref{list:init_task}, which includes historical total volume, city name, county name, population, and freeway information.

\begin{listing}[htp]
\begin{minted}[breaklines,xleftmargin=2em,linenos,fontsize=\scriptsize, escapeinside=!!,breaksymbolleft=]{markdown}
# Task
Construct a time series forecasting model for location `location_id=1201`. Location details: location_id=1201, historical_total_volume=2229867. This location is in Campbell, a city located in Santa Clara County, California. Campbell has a population of approximately 41,700 residents. The location is on freeway SR87-N, which has 3 lanes.

While we could use only data from `location_id=1201`, including data from other locations may improve the model's performance. We request 5 proposals, each suggesting a list of `location_id`s from the neighbors of `location_id=1201`.
\end{minted}
\caption{The task.}
\label{list:init_task}
\end{listing}

Additional guidelines are provided to the agent, as shown in \cref{list:guideline}, detailing what is known about each time series and how sub-datasets should be created.
Sub-datasets are created by selecting time series from the ``neighbors'' of the target time series, defined by similarity in road network distance, local correlation, or geodetic distance.
These ``neighbors'' serve a similar function to retrieved documents in a RAG system~\cite{gao2023retrieval,chang2024main}.

\begin{listing}[htp]
\begin{minted}[breaklines,xleftmargin=2em,linenos,fontsize=\scriptsize, escapeinside=!!,breaksymbolleft=]{markdown}
## Guidelines:
- Ensure each location is selected only once per proposal.
- Utilize the provided neighbor sets based on different criteria (road network, temporal pattern similarity, and geodetic distance).
- Consider the additional details provided for each location, including:
  - Similarity or Distance
  - Historical Total Volume
  - City
  - County
  - Population
  - Freeway
  - Number of Lanes
- Balance the selection of neighbors across different criteria to create diverse and informative proposals.
- Explain the rationale behind each proposal, highlighting how the selected neighbors might contribute to improving the forecasting model.
\end{minted}
\caption{The guidelines.}
\label{list:guideline}
\end{listing}

An example of how neighbors are presented to the agent is shown in \cref{list:neighbor}, providing various details about each time series/location. 
For brevity, we show only the nearest neighbor for location 1201 based on local correlation/pattern similarity.

\begin{listing}[htp]
\begin{minted}[breaklines,xleftmargin=2em,linenos,fontsize=\scriptsize, escapeinside=!!,breaksymbolleft=]{markdown}
## Neighbor Sets:
- Nearest Neighbors Selected Based on Temporal Pattern Similarity.
Neighbors are selected based on the Pearson correlation coefficient between the most similar patterns observed at two locations. This correlation ranges from -1 to 1, indicating the strength and direction of the linear relationship between patterns. This neighbor selection method is particularly valuable because similar patterns across locations suggest that people passing by exhibit comparable behaviors. Consequently, sharing data between these locations when training a model can provide crucial insights into common temporal trends and significantly enhance the model's predictive capabilities. By focusing on temporal similarities rather than geographical proximity, this approach can uncover hidden relationships between seemingly unrelated locations, potentially leading to more nuanced and accurate predictions in various applications such as urban planning, traffic management, or consumer behavior analysis.
  1. location_id=1205, similarity=0.9849, historical_total_volume=2598232. This location is in San Jose, a city located in Santa Clara County, California. San Jose has a population of approximately 969,655 residents. The location is on freeway SR87-N, which has 2 lanes.
  ... (omitted for brevity)
\end{minted}
\caption{The neighbor sets.}
\label{list:neighbor}
\end{listing}

The output format section specifies the format for each proposal, as shown in \cref{list:format}.

\begin{listing}[htp]
\begin{minted}[breaklines,xleftmargin=2em,linenos,fontsize=\scriptsize, escapeinside=!!,breaksymbolleft=]{markdown}
# Output Format
Please output each proposal using the following format:
```
Proposal {proposal_number}
Explanation: {reasoning_behind_the_proposal}
Neighbors: [{location_id_for_neighbor_1}, {location_id_for_neighbor_2}, {location_id_for_neighbor_3}, ..., {location_id_for_last_neighbor}]
```
\end{minted}
\caption{The output format.}
\label{list:format}
\end{listing}

\subsection{Forecasting Module}
\label{sec:forcasting}
The forecasting module trains time series forecasting models using the sub-dataset and evaluates their performance on the validation set.
The \proposed{} framework implements four different time series forecasting models:

\begin{itemize}
    \item Linear~\cite{zeng2023transformers}: A simple yet effective forecasting model.
    \item MLP~\cite{chen2023tsmixer,yeh2024rpmixer}: A natural extension of the linear model with increased learning capability due to additional parameters.
    \item SpraseTSF~\cite{lin2024sparsetsf}: A compact and effective forecasting model.
    \item UltraSTF~\cite{yeh2025compact}: Another compact model with improved cost/learning capability trade-off.
\end{itemize}

To enhance model convergence speed, we first train a foundation model~\cite{yeh2023toward,yeh2025tict,yeh2025treasure,miller2024survey} using all available time series data before user queries.
We then use the sub-dataset proposed by the LLM-agent to fine-tune the foundation model when testing different proposals.
To further refine the data before fine-tuning, we remove the 10\% most anomalous data from the sub-dataset using discord-based anomaly detection methods~\cite{yeh2016matrix,yeh2024matrix}.

\subsection{Proposal Refinement}
\label{sec:refinement}
The proposal refinement agent uses a six-section prompt: 1) objective, 2) background, 3) experiment results, 4) task, 5) additional considerations, and 6) output format.
The overall objective statement and background for the refinement agent are shown in \cref{list:objective}, providing information about the target time series and current best-so-far validation performance.

\begin{listing}[htp]
\begin{minted}[breaklines,xleftmargin=2em,linenos,fontsize=\scriptsize, escapeinside=!!,breaksymbolleft=]{markdown}
# Objective
Develop an improved time series forecasting model for `location_id=1201`, leveraging data from other relevant locations.

# Background
- Target location: `location_id=1201`
- Target location information: location_id=1201, historical_total_volume=2229867. This location is in Campbell, a city located in Santa Clara County, California. Campbell has a population of approximately 41,700 residents. The location is on freeway SR87-N, which has 3 lanes.
- Best performance achieved (Mean Absolute Error): 7.4190
\end{minted}
\caption{The objective.}
\label{list:objective}
\end{listing}

The experiment section lists all sub-dataset setups and their corresponding performance, as shown in \cref{list:experiment}. 
For brevity, we only show the experiment result for one of the proposals. 
Explanations are included to help the agent reason about the main factors behind good performance, and results are ordered based on performance.

\begin{listing}[htp]
\begin{minted}[breaklines,xleftmargin=2em,linenos,fontsize=\scriptsize, escapeinside=!!,breaksymbolleft=]{markdown}
# Previous Experiment Results (Ranked from Best to Worst)
Proposal 1
Explanation: This proposal aims to create a well-rounded selection by integrating one neighbor from each of the three criteria: road network similarity, temporal pattern similarity, and geodetic proximity. Each of these neighbors comes from a different set, ensuring a diverse set of data inputs to help the model generalize better over different types of proximity and similarity.
Neighbors: [1200, 1202, 1204, 1205, 1223]
Performance (Mean Absolute Error): 7.4190
... (omitted for brevity)
\end{minted}
\caption{The experiment results.}
\label{list:experiment}
\end{listing}

Detailed tasks are given to the agent in \cref{list:refine_task}, along with additional considerations to aid in reasoning.
The output format section specifies the format for each proposal, as shown in \cref{list:format}.

\begin{listing}[htp]
\begin{minted}[breaklines,xleftmargin=2em,linenos,fontsize=\scriptsize, escapeinside=!!,breaksymbolleft=]{markdown}
# Task
Based on the experiment results, baseline performance, and best-so-far performance, provide a new set of proposals to further enhance the forecasting model. Each proposal should:
1. Include a list of `location_id`s selected from the neighbors of `location_id=1201`
2. Ensure no duplicate selections within a single proposal
3. Aim to minimize the Mean Absolute Error (MAE)

# Additional Considerations
- Analyze the characteristics of the target location and its neighbors
- Identify patterns in successful proposals from previous experiments
- Explore diverse combinations of locations that may capture various aspects of time series behavior
\end{minted}
\caption{The task.}
\label{list:refine_task}
\end{listing}

\section{Experiment}
This experiment demonstrates the effectiveness of the \proposed{} framework in improving time series forecasting performance and providing valuable insights through explanations.
Key findings show consistent performance improvements across multiple models and the framework's ability to leverage diverse metadata for optimal sub-dataset selection.

\subsection{Dataset and Experiment Setup}
We utilized the LargeST dataset~\cite{liu2023largest}, comprising traffic time series from 8,600 sensors across California.
The dataset includes metadata such as location coordinates, county, freeway name, and number of lanes.
We augmented this metadata with city names, populations, and historical transaction volumes.
Sensor readings were aggregated into 15-minute intervals, resulting in 96 intervals per day over 35,040 time steps.
We split each sub-dataset into training, validation, and test sets using a 6:2:2 ratio.
Our framework was applied to the training and validation sets, with performance reported on the test set.
We tested 60 user queries with randomly selected locations, forecasting the next 12 intervals for each sensor at each timestamp.
Performance was evaluated using Mean Absolute Error (MAE), Root Mean Squared Error (RMSE), and Mean Absolute Percentage Error (MAPE).
GPT-4 Turbo was employed as our LLM agent, generating five proposals per round.

\subsection{Experiment Results}
We applied the \proposed{} framework to four time series forecasting models (see \cref{sec:forcasting}), with results shown in \cref{tab:performance}.
Performance metrics were averaged across all 12 time steps and 60 queries.

\begin{table}[htp]
\caption{
The \proposed{} framework improves performance across all scenarios.
}
\label{tab:performance}
\begin{center}
{\small
\begin{tabular}{lccc}
\shline
Method & MAE & RMSE & MAPE \\
\hline \hline
Linear & 37.31 & 74.01 & 15.12\% \\
Linear+\proposed{} & 35.91 & 72.91 & 14.20\% \\
\hline
\% improvement & 3.77\% & 1.48\% & 6.14\% \\
\hline \hline
MLP & 34.07 & 67.68 & 13.33\% \\
MLP+\proposed{} & 31.26 & 63.34 & 12.34\% \\
\hline
\% improvement & 8.26\% & 6.41\% & 7.42\% \\
\hline \hline
SparseTSF & 37.92 & 74.19 & 16.83\% \\
SparseTSF+\proposed{} & 34.88 & 69.20 & 15.46\% \\
\hline
\% improvement & 8.02\% & 6.73\% & 8.14\% \\
\hline \hline
UltraSTF & 29.77 & 60.92 & 10.26\% \\
UltraSTF+\proposed{} & 28.61 & 57.78 & 9.79\% \\
\hline
\% improvement & 3.91\% & 5.16\% & 4.55\% \\
\shline
\end{tabular}
}
\end{center}
\end{table}

The \proposed{} framework consistently improved performance across all tested methods, with an average improvement of 6\% across all queries, models, and metrics.
This model-agnostic improvement suggests that the framework's data selection process enhances forecast quality regardless of the underlying model.
SparseTSF showed the lowest performance, while UltraSTF combined with the \proposed{} framework achieved the best results.

\begin{figure}[htp]
\centerline{
\includegraphics[width=0.99\linewidth]{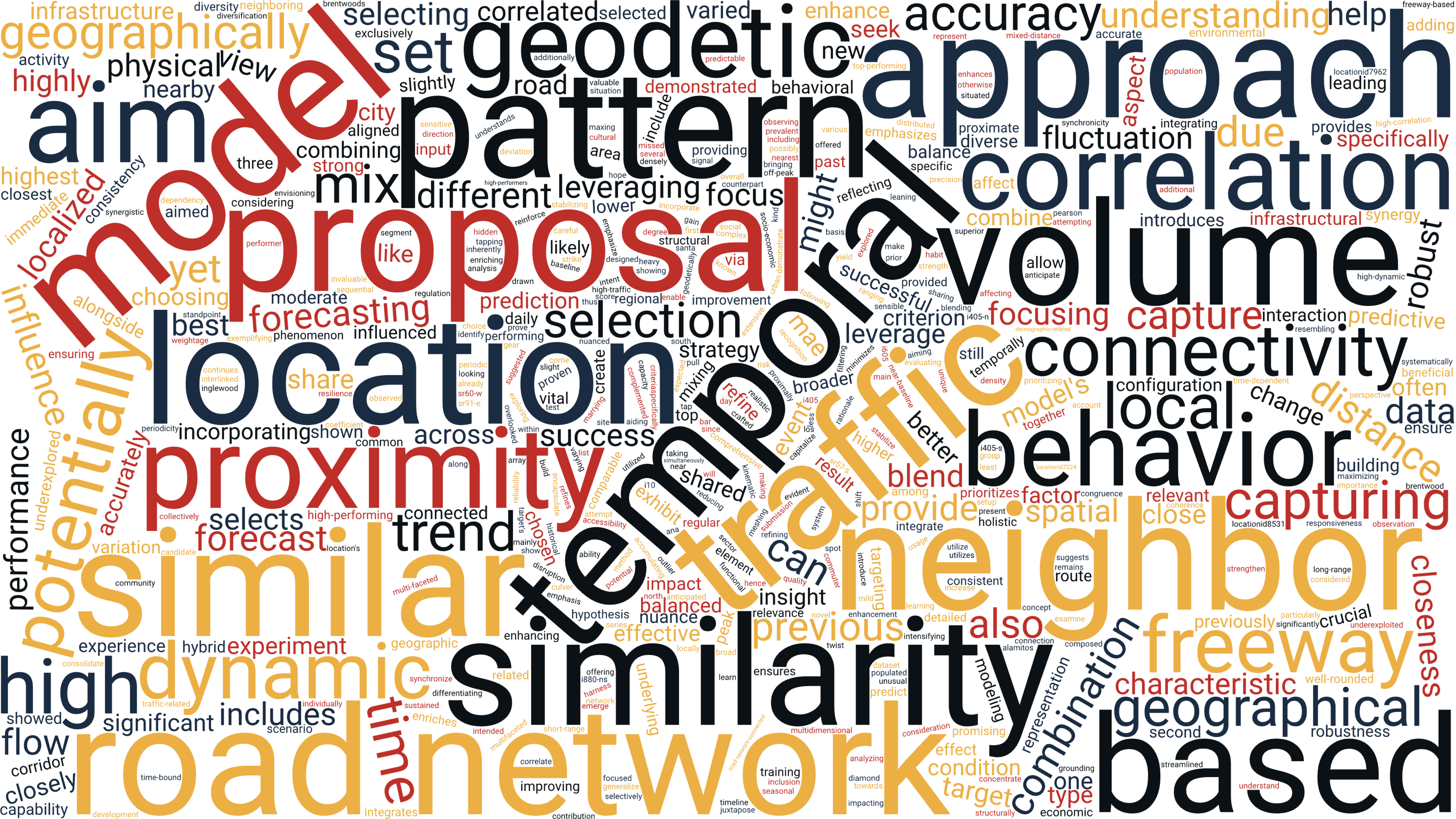}
}
\caption{
Word cloud generated from explanations associated with the best proposal for each user query.
}
\label{fig:wordcloud}
\end{figure}

To understand the agent's proposal generation process, we created a word cloud (\cref{fig:wordcloud}) using explanations associated with the best proposal for all user queries.
The word cloud reveals that terms related to road networks, patterns, and geodetic information have similar prominence, indicating balanced utilization of different neighbor selection criteria across queries.

Next, examining individual query explanations reveals how the framework adapts neighbor selection criteria based on specific query characteristics.
We have selected three proposals that demonstrate how the best proposal for different queries requires neighbors selected using different criteria:

\begin{itemize}
    \item Concentrates on mixing the top-performing neighbors from \textbf{road network similarity} and \textbf{temporal patterns}, which are crucial for understanding structural and sequential traffic volume dynamics.
    \item This proposal focuses on refining the successful aspects of the first proposal from previous experiments, which utilized \textbf{geodetic proximity}... (omitted for brevity).
    \item This proposal includes neighbors selected based on high similarity in \textbf{traffic volume} and \textbf{temporal patterns}, specifically focusing on the \textbf{freeway SR60-W} in \textbf{Diamond Bar}, ... (omitted for brevity).
\end{itemize}

These examples highlight how the agent uses neighbors from different neighbor sets in different proposals, sometimes utilizing specific metadata such as freeway names, cities (e.g., Diamond Bar), or historical traffic volumes to create each proposal.
This observation further motivates the proposed system, as it would be highly labor-intensive for a human to manually analyze the metadata 60 times for the 60 queries.
The \proposed{} framework automates this process, efficiently building the best possible forecast model for each time series within a dataset.

\section{Conclusion}
This paper presents a preliminary study on utilizing LLMs for building time series forecasting models automatically, employing the principles of data-centric AI.
We propose the \proposed{} framework and demonstrate its capability on a traffic volume forecasting dataset, where it improves upon existing systems by 6\%.
Our results suggest that LLMs have significant potential in enhancing time series analysis and forecasting tasks.
For future work, it would be valuable to investigate the performance of the \proposed{} framework when applied to time series from diverse domains and comparing the agents when powered by different LLMs.

\section*{GenAI Usage Disclosure}
The proposed agent is powered by a large language model (LLM). 
Additionally, we use LLMs to proofread and polish our writing.

\bibliographystyle{ACM-Reference-Format}
\bibliography{section/reference}


\end{document}